\documentclass[conference]{IEEEtran}

\usepackage{algorithm}
\usepackage{algorithmicx}
\usepackage{algpseudocode}
\usepackage{amsmath}
\usepackage{subcaption}
\usepackage[dvipsnames,table]{xcolor}
\usepackage[numbers]{natbib}
\usepackage[utf8]{inputenc}
\usepackage{longtable}      % 必须
\usepackage{tabularx}       % 通常需要
\usepackage{booktabs}       % 可选，更好看的表格线
\usepackage{adjustbox}
\usepackage{amssymb}  % amssymb 会自动加载
\usepackage{stfloats}
\usepackage{authblk}
\usepackage[pagebackref,breaklinks,colorlinks]{hyperref}

\title{SAD-GS: Learning Reliable 3D Semantic Gaussian Fields via Dynamic Geo-Semantic Anchoring}

\author[1]{Yufei Zhang}
\author[1]{Chenlu Zhan}
\author[1]{Gaoang Wang}
\author[1]{Hongwei Wang*}

\affil[1]{Zhejiang University}

\begin{document}

\maketitle
\begin{abstract}
Open-vocabulary 3D semantic Gaussian field learning relies on multi-view 2D supervision, whose semantic targets and spatial assignments are often unreliable. Across varying viewpoints, view-dependent features cause semantic identity drift, while propagated tracker masks introduce boundary leakage and identity switches. Directly optimizing against these unreliable 2D targets forces the 3D representation to absorb multi-view contradictions, leading to severe error accumulation. To resolve this limitation, we propose \textbf{SAD-GS}, a framework for learning reliable 3D semantic Gaussian fields via dynamic geo-semantic anchoring. Specifically, \textbf{Semantic Anchor Distillation (SAD)} distills per-view visual embeddings into consensus text anchors to establish a viewpoint-invariant semantic identity. Concurrently, the \textbf{Geo-Semantic Feedback Loop (GSFL)} leverages the evolving 3D field to actively filter tracker anomalies and refine spatial mask assignments via a conservative three-gate update rule. Extensive evaluations on LERF-OVS, 3D-OVS, and Mip-NeRF360 show that \textbf{SAD-GS} consistently achieves the best overall performance in both open-vocabulary localization and semantic segmentation. These comprehensive improvements validate the effectiveness and robustness of dynamic geo-semantic anchoring for reliable 3D semantic Gaussian field learning.

\end{abstract}

\section{Introduction}

Open-vocabulary 3D scene understanding---recognizing and localizing arbitrary objects in reconstructed scenes without predefined categories---is important for robotics, augmented reality, and language-guided embodied AI. In these settings, the key requirement is not only queryability, but reliability: a queried object should remain semantically consistent across viewpoints, clutter, and occlusion. 3D Gaussian Splatting (3DGS)~\cite{kerbl20233dgs} has become a strong substrate for high-fidelity real-time reconstruction, and recent work has extended it with open-vocabulary semantics for queryable 3D perception~\cite{qin2024langsplat, jiang2024open_ov3d, wu2024opengaussian, piekenbrinck2025opensplat3d}. Some methods further specialize this direction to query-conditioned target resolution~\cite{he2025refersplat, junseon2025drsplat}. The common recipe is to extract 2D semantic cues with foundation models and distill them into a 3D field. The central challenge is therefore no longer semantic availability, but supervision reliability under multi-view supervision.

However, existing open-vocabulary 3D pipelines are primarily bottlenecked by supervision reliability rather than the capacity of 3D representations. Standard frameworks typically utilize multi-view render-and-supervise optimization to back-project 2D signals—including SAM-driven masks~\cite{kirillov2023SAM, ravi2024sam2segmentimages, xu2023sampro3d, yin2024sai3d}, CLIP-based region semantics~\cite{radford2021learningclip}, and self-supervised dense vision priors~\cite{caron2021emergingdino, dosovitskiy2020image}—into the 3D field. While sufficient for simplistic environments, this paradigm becomes exceedingly brittle in cluttered settings, where inconsistent 2D supervision is consolidated into the 3D representation and fundamentally degrades downstream localization, segmentation, and query-time reasoning. This trend exposes a core scientific challenge: \emph{how to learn a reliable 3D semantic field from inherently view-variant and noisy 2D semantic supervision?}

This supervision-reliability problem stems from two coupled sources. First, \emph{visual anchor drift}: CLIP image embeddings prioritize global alignment over region-level consistency, causing the same object to produce disparate features across views~\cite{radford2021learningclip, qin2024langsplat, jang2025identityawarelgs}. Second, \emph{tracker label noise}: tracking masks inevitably accumulate identity switches and boundary leakage under occlusions~\cite{ye2024gaussiangrouping, nguyen2024open3dis, cheng2023tracking}. Crucially, these errors compound: feature drift renders semantic identities unidentifiable, while noisy spatial boundaries corrupt the multi-view geometric evidence required to aggregate them. Consequently, standard optimization reinforces rather than averages out these errors across the 3D field. 
Our key insight is that resolving this failure requires treating semantic identity and spatial assignment as two decoupled reliability problems in lifted 2D supervision. The former must be stabilized across views, while the latter must be selectively trusted during 3D optimization. When both signals are treated as fixed pseudo-labels, the 3D field is forced to fit conflicting supervision across views, which accumulates errors in the learned representation.

To this end, we propose \textbf{SAD-GS}, a framework for learning reliable 3D semantic Gaussian fields via dynamic geo-semantic anchoring. Specifically, \textbf{Semantic Anchor Distillation (SAD)} distills fluctuating per-view visual embeddings into consensus text anchors to establish a viewpoint-invariant semantic identity in language space. Concurrently, the \textbf{Geo-Semantic Feedback Loop (GSFL)} leverages the evolving 3D field as a geometric-semantic validator, cross-referencing rendered features, consensus anchors, and local depth consistency to filter tracker anomalies and refine spatial mask assignments via a conservative three-gate update rule. SAD-GS instantiates dynamic geo-semantic anchoring, concurrently stabilizing semantic identities and spatial boundaries during 3D optimization. Our contributions are:
\begin{itemize}
    \item \textbf{Reliability perspective:} We formulate open-vocabulary 3D semantic learning as a supervision-reliability problem under multi-view supervision and identify visual anchor drift and tracker label noise as its two structural failure modes.
    \item \textbf{Supervision-stabilized framework:} We introduce SAD-GS, which stabilizes semantic identity in language space and regulates pseudo-label refinement through geometry-aware feedback.
    \item \textbf{Mechanistic characterization:} We show that semantic anchor stabilization is the primary source of improvement, while geometry-guided feedback provides additional gains but remains bounded by severe overlap and anchor ambiguity.
    \item \textbf{Validation:} We validate SAD-GS across multiple tasks and benchmarks, showing that it consistently improves open-vocabulary 3D localization and segmentation under diverse scene conditions.
\end{itemize}
% Sec.~\ref{sec:related} reviews related work; Sec.~\ref{sec:method} presents the SAD-GS framework; Sec.~\ref{sec:experiments} evaluates all claims experimentally; Sec.~\ref{sec:conclusion} concludes with limitations and future directions.

\section{Related work}
\label{sec:related}
\subsection{2D supervision for 3D semantics}

The extraction of open-vocabulary 3D semantics heavily leverages 2D vision-language foundations as the primary supervisory source. Early paradigms typically distill region-level features from CLIP encoders~\cite{radford2021learningclip}, often augmented by dense visual priors such as DINO~\cite{caron2021emergingdino}, DINOv2~\cite{oquab2024dinov2}, OpenSeg~\cite{ghiasi2022scalingopenseg}, or language-driven 2D segmentation~\cite{li2022languagelseg, wang2024use}, into underlying 3D representations~\cite{qin2024langsplat, liao2024clipgs, zuo2025fmgs}. To further enhance semantic expressiveness, contemporary approaches incorporate descriptive contextual priors generated by modern multi-modal large language models (MLLMs)~\cite{Qwen2.5-VL, chen2024internvl, chen2024expanding, zhu2025internvl3, wang2025internvl3_5} or dense captioning pipelines~\cite{wu2024grit}. While these methods significantly enrich the textual-visual alignment within 3D neural fields, they fundamentally treat supervision as an observation-conditioned proxy. Because the training targets remain anchored to isolated 2D crops, frames, or regional proposals, the resultant supervisory signals inevitably suffer from view-dependent volatility.

A parallel limitation persists regarding geometry-guided mask generation and propagation. Foundation models such as the SAM~\cite{kirillov2023SAM,ravi2024sam2segmentimages}, Grounded-SAM~\cite{ren2024groundedsam}, and VLPart~\cite{sun2023going} provide high-recall spatial proposals, while multi-object tracking frameworks~\cite{cheng2023tracking} establish cross-frame identity association. Open-world pipelines subsequently associate these 2D masks with 3D representations to generate dense pseudo-labels~\cite{tai2024openSAM3D, nguyen2024open3dis, xu2023sampro3d, yin2024sai3d, ying2024omniseg3d}. Although effective for dense scene mapping, existing workflows predominantly optimize the 3D field against these propagated masks as static ground truth, inherently overlooking the tracking drift and accumulated cross-frame errors. In contrast, our work shifts the intervention point upstream: \textbf{SAD} stabilizes the core semantic anchors in the language space to reduce view-dependent semantic drift, while \textbf{GSFL} dynamically evaluates the trustworthiness of propagated labels to selectively filter out observation noise.

\subsection{Open-Vocabulary 3D Scene Understanding}
\noindent\textbf{3D semantic fields.} LERF~\cite{kerr2023lerf} pioneered open-vocabulary 3D querying by lifting language embeddings into volumetric spaces, though its performance remains susceptible to viewpoint-dependent semantic drift. While subsequent non-Gaussian methodologies extended open-world reasoning to point clouds and hierarchical tokens~\cite{ding2023pla, yang2024regionplc, ding2024lowis3d, schult2023mask3d, sun2023superpoint, rozenberszki2024unscene3d, hahn2025cups}, they rarely address the supervision instability inherent in dense, splatting-based pipelines. To fill this gap, LangSplat~\cite{qin2024langsplat} adapted this paradigm to 3D Gaussian Splatting (3DGS) by anchoring CLIP semantics within explicit feature fields. Follow-up 3DGS variants primarily focus on optimizing representation capacity---leveraging contextual reasoning~\cite{wu2024opengaussian}, open-vocabulary transfer~\cite{jiang2024open_ov3d}, foundation-model priors~\cite{zuo2025fmgs}, or structured topological fields~\cite{liao2024clipgs, jang2025identityawarelgs, tian2025ccllgs, li2025_4dlangsplat}. Broader 3DGS research has also improved reconstruction, editing, feed-forward generation, and view-consistent priors~\cite{liu20243dgs, chen2024gaussianeditor, ye2024absgs, chen2024mvsplat, zhou2024diffgs}. Despite these structural advancements, existing workflows overwhelmingly optimize representation scale while naively assuming that the 2D supervision is inherently reliable.

\noindent\textbf{Contrast with Referring 3D Segmentation.} A cognate yet orthogonal line studies referring 3D segmentation. For instance, ReferSplat~\cite{he2025refersplat} specializes in query-conditioned target resolution to untangle complex spatial expressions at inference time. While effective for single-shot expression isolation, such frameworks treat 3D semantics as a transient, instance-centric decoding process tied to specific prompts. In contrast, our framework prioritizes the structural stability and global consistency of the underlying 3D semantic substrate itself \textit{prior to} the application of any query. By shifting from query-dependent decoding to upstream supervision rectification, we establish a persistent, task-agnostic foundation that inherently supports generalized downstream localization and segmentation.

\noindent\textbf{Contrast with Open-Vocabulary Segmentation.} Concurrently, open-vocabulary 3D segmentation frameworks, such as 3D-OVS~\cite{liu2023weakly3dovs}, Open3DIS~\cite{nguyen2024open3dis}, OpenSplat3D~\cite{piekenbrinck2025opensplat3d}, OmniSeg3D~\cite{ying2024omniseg3d}, and SAM-driven 3D segmentation pipelines~\cite{xu2023sampro3d, yin2024sai3d}, push boundaries by aggregating high-recall 2D proposals via multi-view fusion or Gaussian mapping. However, their optimization emphasis rests predominantly on downstream proposal quality, geometric fusion, or feature decoding. In contrast, our work shifts the intervention upstream. Rather than mitigating multi-view noise during 3D amalgamation, we selectively regulate the trustworthiness of individual 2D supervisory labels prior to back-propagation.

\section{Methodology}
\label{sec:method}
\begin{figure*}[ht]
    \centering
    \includegraphics[width=1\linewidth]{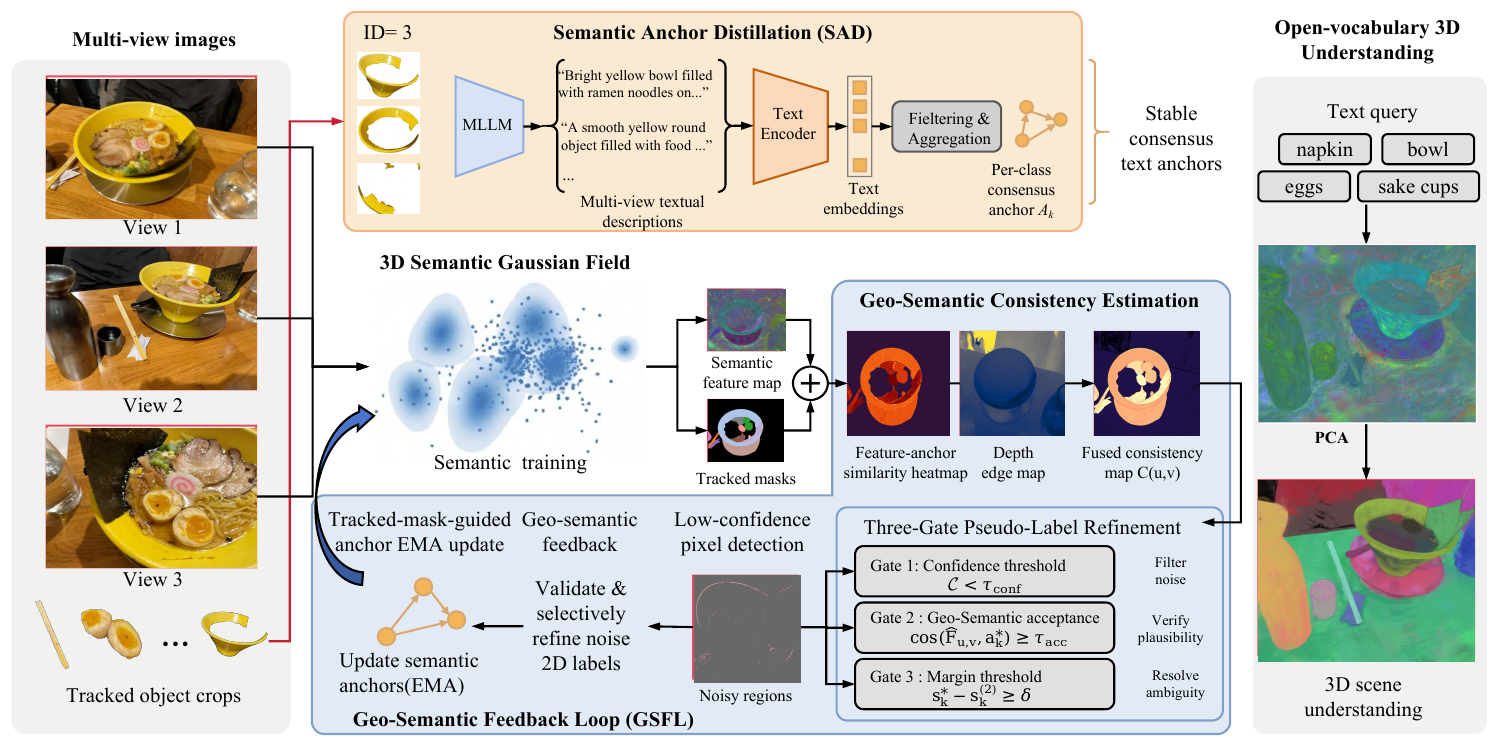}
    \caption{\textbf{Overview of SAD-GS.} SAD-GS learns open-vocabulary 3D semantic fields from multi-view supervision by stabilizing its semantic target and refining its propagated labels. SAD converts multi-view semantic descriptions into stable consensus anchors, and GSFL uses geo-semantic feedback to refine unreliable tracked labels during training.}
    \label{fig:overview}
\end{figure*}

SAD-GS learns open-vocabulary 3D semantic Gaussian fields through dynamic geo-semantic anchoring. As shown in Fig.~\ref{fig:overview}, SAD establishes consensus semantic anchors from multi-view object descriptions, while GSFL uses the evolving 3D field to refine unreliable propagated labels during semantic optimization.

\subsection{Semantic Gaussian Splatting}
\label{sec:preliminaries}
Our goal is to learn a reliable 3D semantic field from unreliable multi-view supervision. We build on 3D Gaussian Splatting (3DGS), which represents a scene with anisotropic Gaussian primitives parameterized by position $\mu_i \in \mathbb{R}^3$, covariance $\Sigma_i$, opacity $\alpha_i$, color coefficients $c_i$, and a semantic feature vector $f_i \in \mathbb{R}^d$ ($d=16$ in practice). Projecting these primitives onto the image plane yields both color $C$ and semantic features $S$ through the same differentiable rendering weights:
\begin{equation}
\small
    C = \sum_{i=1}^{M} c_i T_i \alpha'_i, \quad S = \sum_{i=1}^{M} f_i T_i \alpha'_i, \quad T_i = \prod_{j=1}^{i-1} (1 - \alpha'_j)
    \label{eq:rendering}
\end{equation}
where $\alpha'_i$ denotes the effective 2D opacity and $T_i$ is the accumulated transmittance. Conventional open-vocabulary 3D pipelines optimize the rendered semantic map $S$ against 2D pseudo-labels $S_{\text{2D}}$ from foundation models via
\begin{equation}
    \mathcal{L}_{\text{sem}} = \mathcal{D}(S, S_{\text{2D}}),
\end{equation}
where $\mathcal{D}(\cdot, \cdot)$ is the discrepancy function, typically instantiated as an $L_1$ distance\cite{qin2024langsplat} or cosine similarity\cite{qu2024goi}. While \textbf{SAD-GS} inherits this foundational formulation, it departs from prior practices by substituting these volatile 2D targets with consensus text anchors and dynamically refining noisy spatial assignments during optimization.

\subsection{Semantic Anchor Distillation (SAD)}
\label{sec:sad}
Directly distilling 2D crop features into 3D makes the supervision target viewpoint-dependent: the same object can produce different features under pose, lighting, and occlusion changes. SAD replaces these volatile visual targets with a stable language-space identity target. Concretely, it builds a per-class consensus anchor $\mathbf{a}_k$ from multi-view object descriptions, projects rendered 3D semantic features into CLIP space, and aligns them to the corresponding anchor.

\noindent\textbf{Consensus Anchor Generation.} 
For each tracked object $k$, we sample representative views $\mathcal{V}_k$, extract object crops $\mathcal{I}_{k,j}$, and use Qwen3-VL~\cite{Qwen3-VL} to generate descriptions $d_{k,j}$. These descriptions are mapped to CLIP text embeddings $z_{k,j} = E_{\text{text}}(d_{k,j})$ and aggregated into a \textit{Semantic Consensus Anchor} $\mathbf{a}_k$:
\begin{equation}
    \mathbf{a}_k = \text{Normalize}\left( \sum_{j=1}^{n} w_{k,j} z_{k,j} \right),
\end{equation}
where $w_{k,j}$ filters out description outliers according to the cosine similarity to the initial mean $\bar{z}_k = \frac{1}{n} \sum_{j=1}^n z_{k,j}$:
\begin{equation}
    w_{k,j} = \mathbb{I}\left( \cos(z_{k,j}, \bar{z}_k) > \tau \right),
\end{equation}
where $\mathbb{I}(\cdot)$ is the indicator function and $\tau$ is a robustness threshold. This yields a stable anchor that summarizes the shared identity across views.

\noindent\textbf{Projection Layer.}
Because 3DGS stores a compact $d{=}16$-dimensional semantic feature per Gaussian, we use a learned projection head $\phi: \mathbb{R}^{16} \to \mathbb{R}^{512}$ ($16 \to 128 \to 512$ with ReLU) to map rendered features $\mathbf{F} \in \mathbb{R}^{16 \times H \times W}$ into the CLIP space shared by the anchors.

\noindent\textbf{Distillation Loss.} 
During training, the projected rendered feature $\hat{\mathbf{F}}_{u,v} = \phi(\mathbf{F}_{u,v}) \in \mathbb{R}^{512}$ for each pixel $(u,v)$ is aligned with the anchor $\mathbf{a}_{M_\text{2D}(u,v)}$ of its assigned class via a combined distillation loss:
\begin{equation}
    \label{eq:sad_loss}
    \begin{aligned}
    \mathcal{L}_{\text{sad}}
    = \frac{1}{|\Omega|}\sum_{(u,v) \in \Omega}
    \bigg( &\bigl\| \hat{\mathbf{F}}_{u,v} - \mathbf{a}_{M_\text{2D}(u,v)} \bigr\|_2 \\
    &+ \frac{1}{2}\big( 1 - \cos\bigl(\hat{\mathbf{F}}_{u,v}, \mathbf{a}_{M_\text{2D}(u,v)}\bigr) \big) \bigg),
    \end{aligned}
\end{equation}
where $\Omega$ is the set of foreground pixels and $M_\text{2D}(u,v)$ is the tracker-assigned class ID. The L2 term enforces absolute feature alignment, while the cosine term enforces directional alignment in CLIP space. In the full SAD-GS model, $\mathcal{L}_{\text{sad}}$ uses uniform pixel weights; the consistency score $C(u,v)$ is used only for pseudo-label refinement and anchor updates.

\begin{figure*}[ht]
    \centering
    \includegraphics[width=\linewidth]{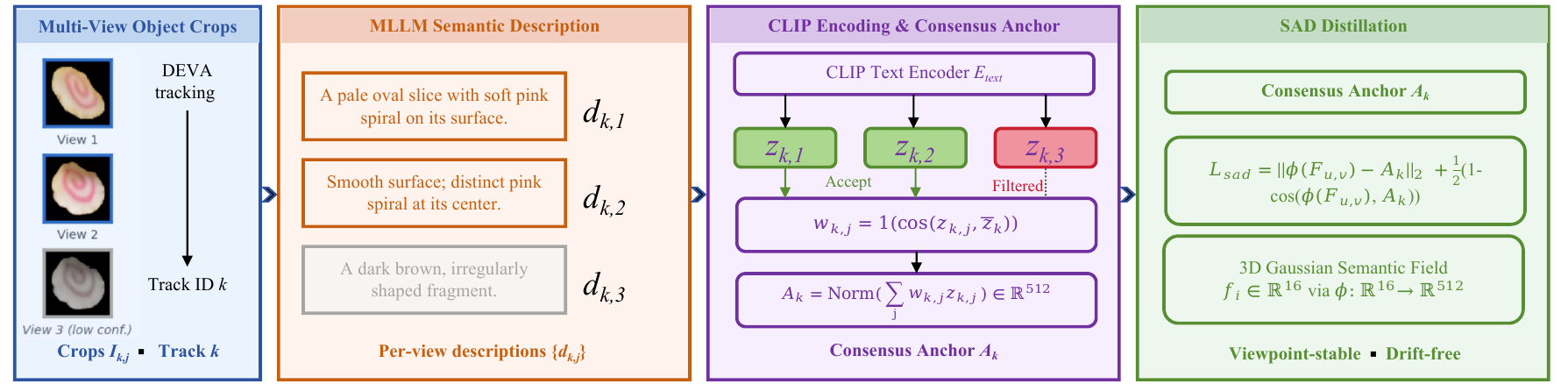}
    \caption{\textbf{Semantic Anchor Distillation (SAD).}
    Multi-view tracked object crops are described by Qwen3-VL, filtered in CLIP space, and aggregated into a consensus anchor for each object. The resulting viewpoint-stable anchors supervise the 3D Gaussian semantic features through $\mathcal{L}_{\text{sad}}$.}
    \label{fig:sad_pipeline}
\end{figure*}

\subsection{Geo-Semantic Feedback Loop (GSFL)}
\label{sec:gsfl}
SAD stabilizes semantic identity, but it does not guarantee that the tracked 2D masks are correct. We therefore introduce the \textbf{Geo-Semantic Feedback Loop (GSFL)} to address noisy propagated masks. We use DEVA~\cite{cheng2023tracking} to obtain initial tracked masks. Although temporally coherent, these masks can still suffer from boundary leakage and ID drift. By contrast, the evolving 3D Gaussian field aggregates evidence across views and becomes geometrically more reliable. GSFL uses this complementary signal as a geometric validator: it identifies pixels whose tracked labels conflict with the current 3D field and refines only these low-confidence regions.

\noindent\textbf{Mechanistic Failures of Naive Self-Refinement.} 
A naive alternative is to directly overwrite ambiguous 2D labels with rendered predictions. In practice, this unconstrained self-refinement fails for two reasons:
\begin{itemize}
    \item \textbf{Spatial overlap:} In nested or co-located regions, aggressive relabeling allows dominant classes to absorb nearby pixels and trigger semantic collapse.
    \item \textbf{Fine-grained ambiguity:} In scenes with visually similar instances, multiple anchors can satisfy the same confidence threshold, so naive relabeling accepts many wrong replacements.
\end{itemize}
These failure modes motivate the three-gate design of GSFL. Rather than relabeling whenever a new prediction looks plausible, GSFL verifies when relabeling is needed, whether the candidate label is reliable, and whether it is clearly better than its nearest competitor.

\noindent\textbf{Geo-Semantic Consistency Score.}
At each training step, we render the per-pixel raw feature map $\mathbf{F} \in \mathbb{R}^{16 \times H \times W}$ and the depth map $D \in \mathbb{R}^{H \times W}$ from the current 3D Gaussian field. We apply the projection head $\phi$ (Sec.~\ref{sec:sad}) to obtain CLIP-space features $\hat{\mathbf{F}}_{u,v} = \phi(\mathbf{F}_{u,v}) \in \mathbb{R}^{512}$. Let $\mathbf{a}_{M_\text{2D}(u,v)} \in \mathbb{R}^{512}$ be the semantic anchor of the class assigned to pixel $(u,v)$ by the 2D tracker mask $M_{\text{2D}}$. The \textit{geo-semantic consistency score} $C(u,v)$ combines feature-anchor agreement with local geometric smoothness:
\begin{equation}
    C(u,v) = \underbrace{\frac{1 + \cos\!\left(\hat{\mathbf{F}}_{u,v},\, \mathbf{a}_{M_\text{2D}(u,v)}\right)}{2}}_{\text{semantic term}} \cdot \underbrace{\exp\!\left(-\lambda \,|\nabla D|_{u,v}\right)}_{\text{geometric term}},
    \label{eq:gsfl_score}
\end{equation}
where $|\nabla D|_{u,v}$ is the local depth gradient magnitude (approximated via finite differences) and $\lambda{=}0.1$ controls the penalty for depth discontinuities. This single-view depth gradient serves as an efficient proxy for multi-view geometric uncertainty: depth edges in a rasterized Gaussian field co-occur with ambiguous surface ownership, making them reliable indicators of label noise. The semantic term is high when the rendered feature agrees with the tracker's class anchor; the geometric term is low at depth edges, where 3D boundaries are uncertain.

\noindent\textbf{Consistency-Guided Label Quality.}
The consistency score $C(u,v)$ is the entry criterion for GSFL. Pixels with $C(u,v) < \tau_{\text{conf}}$ are treated as low-confidence and passed to pseudo-label refinement, while high-confidence pixels keep their tracked labels. GSFL activates only after the semantic field has warmed up: it starts after $T_{\text{warmup}}=5{,}000$ semantic iterations (i.e., iteration~35{,}000 in total training), and $\tau_{\text{conf}}$ is linearly ramped from $\tau_0=0.50$ to $\tau_1=0.65$ over 10{,}000 iterations to favor safe correction early and stricter filtering once the semantic field becomes more reliable.

\noindent\textbf{Three-Gate Pseudo-Label Refinement.}
\label{sec:margin_gate}
GSFL refines labels conservatively. For each low-confidence foreground pixel, it first checks whether the pixel should even be considered for relabeling (Gate~1: low consistency). It then queries the anchor space for the best alternative label $k^{*} = \arg\max_k \cos(\hat{\mathbf{F}}_{u,v}, \mathbf{a}_k)$. This candidate is accepted only if it passes two additional tests:
\begin{equation}
    \underbrace{\cos(\hat{\mathbf{F}}_{u,v}, \mathbf{a}_{k^*}) \geq \tau_{\text{acc}}}_{\text{Gate 2: acceptance}} \quad \wedge \quad \underbrace{s_{k^*} - s_{k^{(2)}} \geq \delta}_{\text{Gate 3: margin}},
    \label{eq:margin_gate}
\end{equation}
where $s_{k^{(2)}}$ is the second-highest anchor similarity. We set $\tau_{\text{acc}}{=}0.60$ and $\delta{=}0.05$ as conservative thresholds to favor safe relabeling. 

The three gates address three different failure sources in pseudo-label correction. Gate~1 restricts the correction scope by selecting only pixels with tracking noise (low consistency), which avoids rewriting correctly labeled pixels. Gate~2 prevents weak-evidence errors; a new label is accepted only when its anchor similarity passes an absolute threshold, avoiding forced corrections under occlusion or unstable views. Gate~3 resolves fine-grained semantic ambiguity; when multiple anchors are plausible, the new label is accepted only if the top candidate is clearly separated from the runner-up, preventing ambiguous relabeling among similar classes. Algorithm~\ref{alg:three_gate} summarizes the decision.

\begin{algorithm}[ht]
\small
\caption{Three-Gate Pseudo-Label Refinement (per training view, GSFL-active phase)}
\label{alg:three_gate}
\begin{algorithmic}[1]
\Require Rendered features $\hat{\mathbf{F}}$, consistency map $C$, tracker mask $M_{\text{2D}}$, anchors $\{\mathbf{a}_k\}_{k=1}^K$
\Require Thresholds: $\tau_{\text{conf}} \in [0.50,\,0.65]$ (linearly ramped), $\tau_{\text{acc}}{=}0.60$, $\delta{=}0.05$
\Ensure Refined mask $M'$ \Comment{called only when iter $\geq T_{\text{warmup}}$}
\State $M' \leftarrow M_{\text{2D}}$ \Comment{initialise from tracker labels}
\For{each foreground pixel $(u,v)$}
    \If{$C(u,v) < \tau_{\text{conf}}$} \Comment{Gate~1: pixel is low-confidence}
        \State $\mathbf{s} \leftarrow \bigl[\cos(\hat{\mathbf{F}}_{u,v}, \mathbf{a}_k)\bigr]_{k=1}^K$
        \State $k^* \leftarrow \arg\max_k s_k$;\quad $s^{(2)} \leftarrow \text{second-largest}(\mathbf{s})$
        \If{$s_{k^*} \geq \tau_{\text{acc}}$} \Comment{Gate~2: absolute acceptance}
            \If{$s_{k^*} - s^{(2)} \geq \delta$} \Comment{Gate~3: margin gate}
                \State $M'(u,v) \leftarrow k^*$ \Comment{accept refined label}
            \EndIf
        \EndIf
    \EndIf
\EndFor
\State \Return $M'$
\end{algorithmic}
\end{algorithm}

\noindent\textbf{Tracked-Mask-Guided Anchor Update.}
To keep the semantic anchors aligned with the evolving 3D feature distribution, we update them every $\Delta_{\text{anc}}=500$ iterations using an EMA driven by the tracked masks rather than model predictions:
\begin{align}
    \mathbf{a}_k \;\leftarrow\; \operatorname{norm}\!\left(m\,\mathbf{a}_k + (1-m)\,\bar{\mathbf{F}}_k^{\text{mask}}\right), \\
    \bar{\mathbf{F}}_k^{\text{mask}} = \operatorname{norm}\!\!\left(\frac{1}{|\mathcal{P}_k|}\sum_{(u,v)\in\mathcal{P}_k}\hat{\mathbf{F}}_{u,v}\right),
    \label{eq:anchor_ema}
\end{align}
where $\mathcal{P}_k = \{(u,v) \mid M_{\text{2D}}(u,v)=k\}$ is the set of pixels assigned to class $k$ by the tracked mask, and $m=0.999$ is the EMA momentum. This avoids circular self-prediction updates.

\noindent\textbf{Training Objective.}
SAD-GS optimizes the semantic feature vectors $\{f_i\}$ and projection head $\phi$ while freezing the geometric parameters from the pre-trained geometry stage. The full training objective is
\begin{equation}
    \mathcal{L}_{\text{total}} = \mathcal{L}_{\text{sad}}(M'),
    \label{eq:total_loss}
\end{equation}
where $M'$ denotes the effective supervision mask: $M'{=}M_{\text{2D}}$ during warm-up and $M'$ is the refined mask from Algorithm~\ref{alg:three_gate} after GSFL activation. Anchor updates (Eq.~\ref{eq:anchor_ema}) are applied in-place and contribute no gradient to $\mathcal{L}_{\text{total}}$.

\subsection{Open-vocabulary Querying}
After training, the 3D semantic field can be queried directly with natural language prompts. Each position $(u,v)$ yields a rendered semantic feature $\hat{\mathbf{F}}_{u,v} \in \mathbb{R}^{512}$ in the CLIP feature space. Given a query phrase $q$, we extract its text embedding $\mathbf{e}_q = E_{\text{text}}(q)$ and introduce four canonical background embeddings $\{\mathbf{e}_{\text{can}}^i\}$ (\textit{object, things, stuff, texture}) following~\cite{qin2024langsplat}. The final query relevancy score $r(u,v)$ is computed using a softmax formulation scaled by a decoding temperature $T$:
\begin{equation}
    r(u,v) = \min_{i} \frac{\exp\left(\hat{\mathbf{F}}_{u,v}\cdot \mathbf{e}_q / T\right)}{\exp\left(\hat{\mathbf{F}}_{u,v}\cdot \mathbf{e}_q / T\right) + \exp\left(\hat{\mathbf{F}}_{u,v}\cdot \mathbf{e}_{\text{can}}^i / T\right)} .
\end{equation}
For \textbf{3D localization}, we retrieve the 3D position of the Gaussian point with the highest rendered relevancy. For \textbf{open-vocabulary 3D segmentation}, we extract the target masks by filtering the relevancy map with a foreground threshold $f_{\text{fg}}$, retaining points where $r(u,v) > f_{\text{fg}}$. Here, $T$ controls the sharpness of the query-background discrimination, while $f_{\text{fg}}$ determines the boundary conservatism of the decoded foreground mask.

\section{Experiments}
\label{sec:experiments}
\subsection{Experimental Setup}
\label{sec:experiments_setup}

\noindent\textbf{Datasets and Metrics.} 
We evaluate whether supervision stabilization leads to more reliable open-vocabulary 3D semantics across three benchmarks with complementary scene characteristics and evaluation protocols:
\begin{itemize}
    \item \textbf{LERF-OVS}~\cite{kerr2023lerf} contains cluttered real-world scenes with heavy object overlap (e.g., \textit{ramen}). We evaluate both open-vocabulary 3D localization with localization accuracy and open-vocabulary 3D segmentation with mIoU.
    \item \textbf{3D-OVS}~\cite{liu2023weakly3dovs} features relatively isolated long-tail object categories captured from diverse viewpoints and backgrounds. We use this benchmark to evaluate open-vocabulary 3D segmentation under weak supervision, reported as overall mIoU.
    \item \textbf{Mip-NeRF360}~\cite{barron2021mipnerf} covers large-scale unbounded indoor and outdoor scenes with complex geometry. Using labels annotated by GAGS~\cite{guo2024semanticgs}, we report mean IoU to assess cross-scene generalization.
\end{itemize}

\noindent\textbf{Implementation Details.}
Our framework is built on standard 3DGS~\cite{kerbl20233dgs}. For each scene, we first optimize the geometric field under RGB supervision for 30,000 iterations. We then freeze all geometric parameters and optimize the 16-dimensional semantic feature field and projection head for another 20,000 iterations. For 2D prior generation, we use ViT-H SAM~\cite{kirillov2023SAM} for mask proposals and Qwen3-VL~\cite{Qwen3-VL} for object-centric descriptions. Experiments are run on NVIDIA RTX PRO 6000 GPUs. Training takes about 2 hours per scene, including roughly 30 minutes for geometry, 90 minutes for semantic training, and about 10 minutes for offline anchor generation.

\begin{figure}[ht]
    \centering
    \includegraphics[width=1\linewidth]{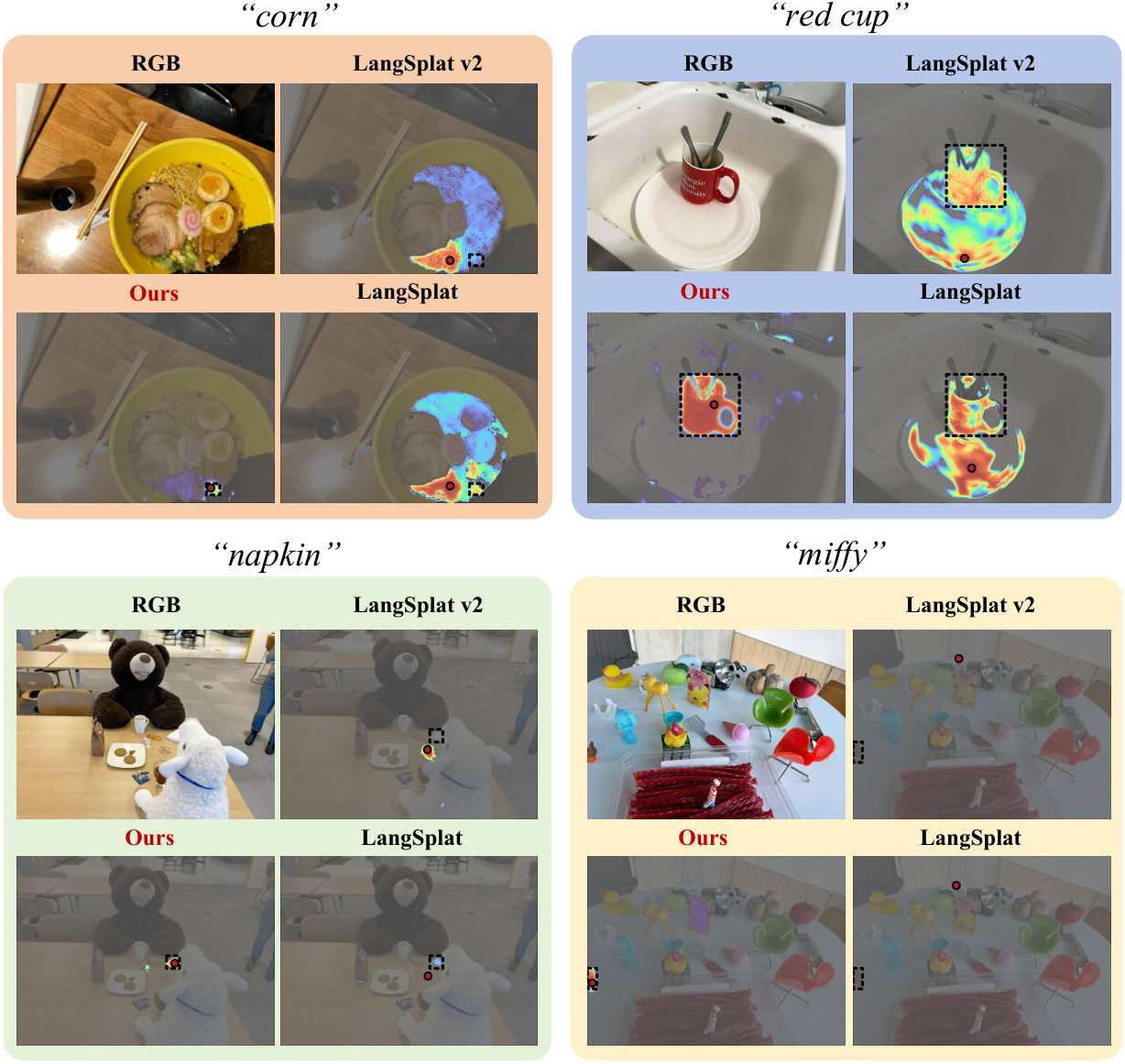}
\caption{\textbf{Qualitative localization comparison on LERF-OVS.} SAD-GS yields sharper query responses and more accurate target localization than prior methods.}
    \label{fig:loc_vis}
\end{figure}
\subsection{3D Object Localization}
Table~\ref{tab:loc} reports the quantitative localization results. \textbf{SAD-GS} achieves the best overall accuracy on LERF-OVS. The largest gain appears on the challenging \textit{ramen} scene, where accuracy improves from $74.7\%$ with LangSplat--v2~\cite{li2026langsplatv2} to $85.8\%$. This improvement is consistent with the role of SAD: stable text-consensus anchors reduce viewpoint-dependent semantic drift and make query-conditioned localization more reliable.

\begin{table}[ht]
\centering
\caption{Quantitative comparisons of open-vocabulary 3D object localization. We report the mean accuracy $(\%)$.}
\label{tab:loc}
\resizebox{1\linewidth}{!}{
\begin{tabular}{lccccc}
\toprule
\textbf{Method} & Ramen & Teatime & Kitchen & Figurines & Overall \\
\midrule
LERF~\cite{kerr2023lerf} & $62.0$ & $84.8$ & $72.7$ & $75.0$ & $73.6$ \\
GS-Grouping~\cite{ye2024gaussiangrouping} & $32.4$ & $69.5$ & $50.0$ & $44.6$ & $49.1$ \\
OpenGaussian~\cite{wu2024opengaussian} & $29.4$ & $61.6$ & $31.9$ & $61.1$ & $46.0$ \\
LangSplat~\cite{qin2024langsplat} & $73.2$ & $88.1$ & $95.5$ & $80.4$ & $84.3$ \\
GOI~\cite{qu2024goi} & $56.3$ & $67.8$ & $68.2$ & $44.6$ & $59.2$ \\
LangSplat--v2~\cite{li2026langsplatv2} & $74.7$ & $93.2$ & $86.4$ & $82.1$ & $84.1$ \\ \midrule
Ours & $\mathbf{85.8}$ & $\mathbf{89.3}$ & $\mathbf{96.5}$ & $\mathbf{83.2}$ & $\mathbf{88.7}$ \\
\bottomrule
\end{tabular}}
\end{table}

\begin{table}[ht]
\centering
\caption{Component ablation for 3D localization on \textit{figurines} and \textit{ramen}. $\Delta$: improvement over LangSplat--v2~\cite{li2026langsplatv2}.}
\label{tab:abl-loc}
\resizebox{1\linewidth}{!}{
\begin{tabular}{lc|cc|cc}
\toprule
SAD & GSFL &  Figurines & & Ramen \\
\midrule
 — & — & Acc (\%) & $\Delta$ & Acc (\%) & $\Delta$ \\
\midrule
 — & \checkmark & 81.3 & +0.9 & $76.0$ & $+1.3$\\
\checkmark & — & 81.9 & +1.5 & $81.6$ & $+6.9$\\
 \checkmark & \checkmark & \textbf{83.2} & \textbf{+1.9} & \textbf{85.8} &\textbf{+11.1}  \\
\bottomrule
\end{tabular}
}
\end{table}

Table~\ref{tab:abl-loc} shows that both SAD and GSFL consistently improve open-vocabulary 3D localization, with the full \textbf{SAD-GS} model achieving the best results. Compared with other baselines, SAD successfully reduces multi-view query drift, while GSFL further sharpens the localization response in regions with heavy spatial overlap or visual confusion. These results demonstrate that our model effectively stabilizes the 3D language field for precise object localization. Fig.~\ref{fig:loc_vis} visually confirms this quantitative trend. 

\begin{figure*}[ht]
    \centering
    \includegraphics[width=0.9\linewidth]{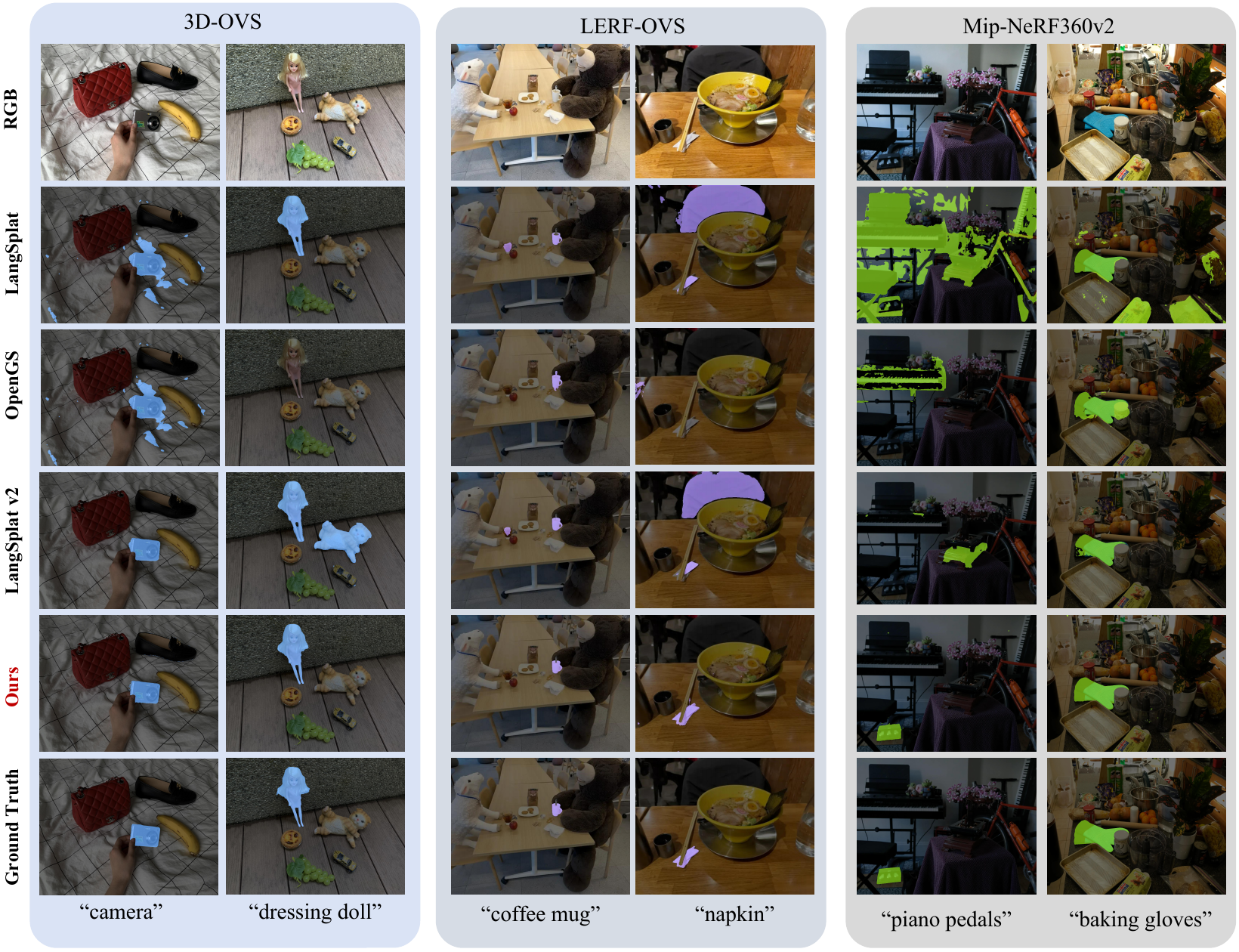}
\caption{\textbf{Cross-dataset qualitative segmentation comparison.} Across LERF-OVS, Mip-NeRF360, and 3D-OVS, SAD-GS produces cleaner query masks with less semantic leakage than prior methods.}
    \label{fig:seg_querymask}
\end{figure*}

\subsection{3D Semantic segmentation.}
\noindent\textbf{Results on the LERF-OVS Dataset.}
Table~\ref{tab:seg1} shows that \textbf{SAD-GS} achieves the best overall performance on LERF-OVS. Compared with LangSplat--v2, it improves the overall mIoU from $59.9\%$ to $68.8\%$. This result shows that supervision stabilization improves open-vocabulary 3D semantic segmentation in cluttered scenes.

\begin{table}[ht]
\centering
\caption{Quantitative 3D semantic segmentation results on LERF-OVS, reported as mean IoU ($\%$).}
\label{tab:seg1}
\resizebox{1\linewidth}{!}{
\begin{tabular}{lccccc}
\toprule
\textbf{Method} & Ramen & Teatime & Kitchen & Figurines & Overall \\
\midrule
GS-Grouping~\cite{ye2024gaussiangrouping} & $26.4$ & $54.0$ & $31.3$ & $34.6$ & $36.6$ \\
Refersplat~\cite{he2025refersplat} & $55.1$ & $50.1$ & $67.5$ & $48.9$ & $55.4$ \\
OpenGaussian~\cite{wu2024opengaussian} & $31.0$ & $60.4$ & $39.3$ & $22.7$ & $38.4$ \\
LangSplat~\cite{qin2024langsplat} & $51.2$ & $65.1$ & $44.5$ & $44.7$ & $51.4$ \\
LangSplat--v2~\cite{li2026langsplatv2} & $51.8$ & $72.2$ & $59.1$ & $56.4$ & $59.9$ \\
SparseLGS~\cite{chensparselgs} & $55.2$ & $40.7$ & $47.3$ & $22.9$ & $41.5$ \\ \midrule
SAD-only$^\star$ & 72.5 & 67.2 & 63.7 & 46.3 & 63.7 \\
    \textbf{SAD-GS} & \textbf{77.9} & \textbf{72.2} & \textbf{68.4} & \textbf{57.6} & \textbf{68.8}\\
\bottomrule
\end{tabular}%
}
\end{table}

\noindent\textbf{Results on the 3D-OVS Dataset.} 
Table~\ref{tab:seg2} shows the quantitative comparisons on the 3D-OVS dataset. We see that \textbf{SAD-GS} achieves the highest overall mIoU score of $\mathbf{96.0\%}$, outperforming LangSplat by $2.6\%$ and the SOTA LangSplatv2 by $1.4\%$. This consistent improvement further demonstrates that our model effectively builds more stable and view-invariant 3D semantic fields.

\begin{table}[ht]
\centering
\caption{Quantitative 3D semantic segmentation results on 3D-OVS full scenes, reported as mean IoU (\%).}
\label{tab:seg2}
\resizebox{1\linewidth}{!}{%
\begin{tabular}{lcccccc}
\toprule
\textbf{Method} & Bed & Bench & Room & Sofa & Lawn & Overall \\
\midrule
LERF\cite{kerr2023lerf} & $73.5$ & $53.2$ & $46.6$ & $27.0$ & $73.7$ & $54.8$ \\
Refersplat\cite{he2025refersplat} & $96.3$ & $92.1$ & $96.7$ & $95.0$ & $88.9$ & $93.8$ \\
OpenGaussian\cite{wu2024opengaussian} & $91.2$ & $88.3$ & $93.5$ & $89.4$ & $93.3$ & $91.1$ \\
3D-OVS~\cite{yu2024mip} & $89.5$ & $89.3$ & $92.8$ & $74.0$ & $88.2$ & $86.8$ \\
LangSplat\cite{qin2024langsplat} & $92.5$ & $94.2$ & $94.1$ & $90.0$ & $96.1$ & $93.4$ \\
LangSplat--v2\cite{li2026langsplatv2} & $93.0$ & $94.9$ & $96.1$ & $92.3$ & $96.6$ & $94.6$ \\ \midrule
Ours & $\mathbf{98.4}$ & $\mathbf{97.7}$ & $\mathbf{96.6}$ & $\mathbf{92.0}$ & $\mathbf{95.2}$ & $\mathbf{96.0}$ \\ 
\bottomrule
\end{tabular}%
}
\end{table}

\noindent\textbf{Results on the Mip-NeRF360 Dataset.} 
Table~\ref{tab:seg3} further validates cross-scene generalization on the more complex Mip-NeRF360 scenes. SAD-GS reaches $\mathbf{75.4\%}$ overall mIoU, exceeding LangSplat--v2 by $6.0\%$, with the largest gain on \textit{bonsai}.

\noindent\textbf{Qualitative Results.} Fig.~\ref{fig:seg_querymask} shows the same trend across all three datasets. Compared with prior methods, SAD-GS produces more compact query masks with less spillover to nearby regions, consistent with the quantitative gains in Tables~\ref{tab:seg1}--\ref{tab:seg3}.

\begin{figure*}[ht]
  \centering
  \begin{minipage}[ht]{0.48\linewidth}
    \centering
    \includegraphics[width=\linewidth]{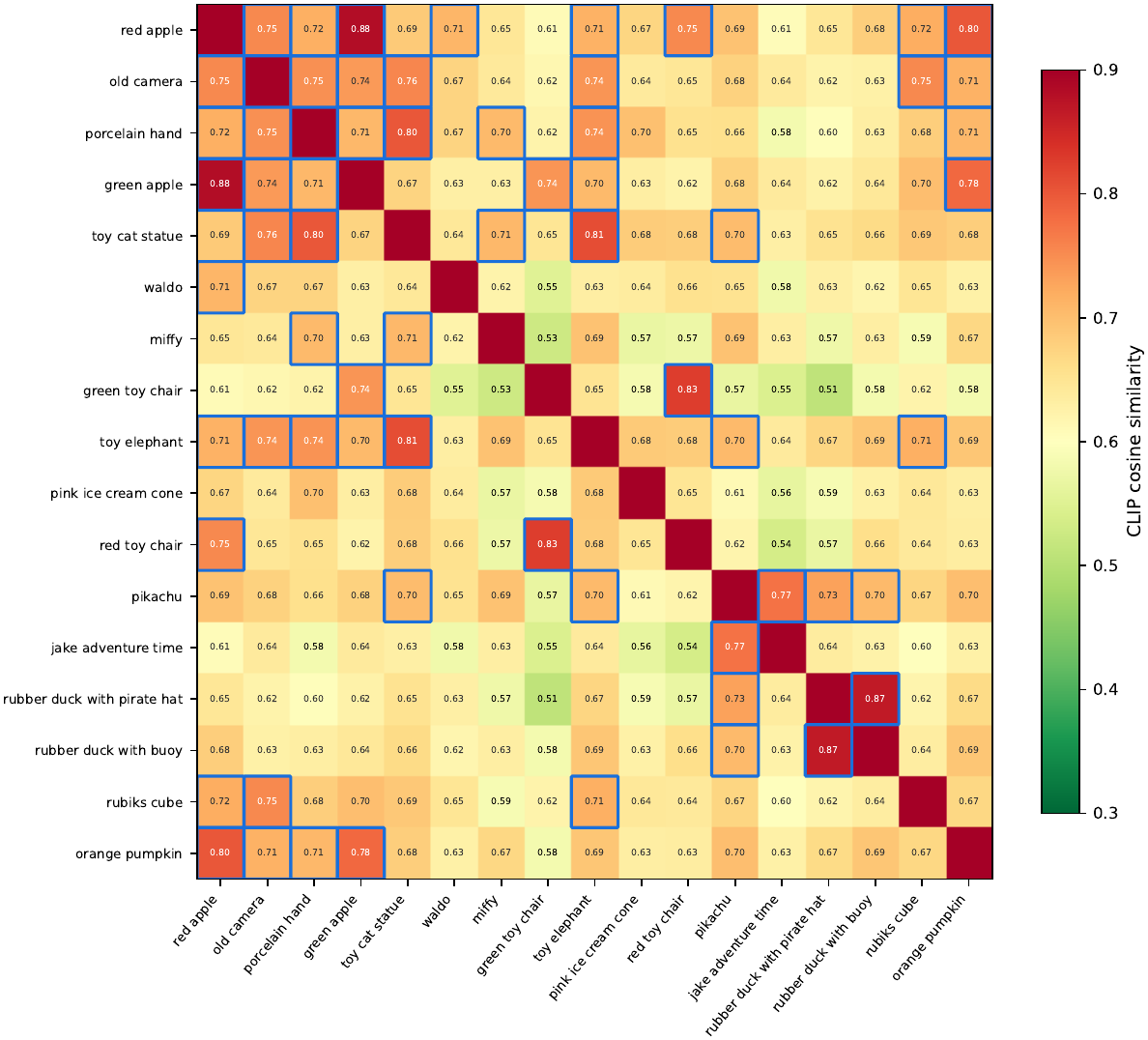}
    \subcaption{\textit{figurines}: global confusion.}
    \label{fig:sim-figurines}
  \end{minipage}\hfill
  \begin{minipage}[ht]{0.48\linewidth}
    \centering
    \includegraphics[width=\linewidth]{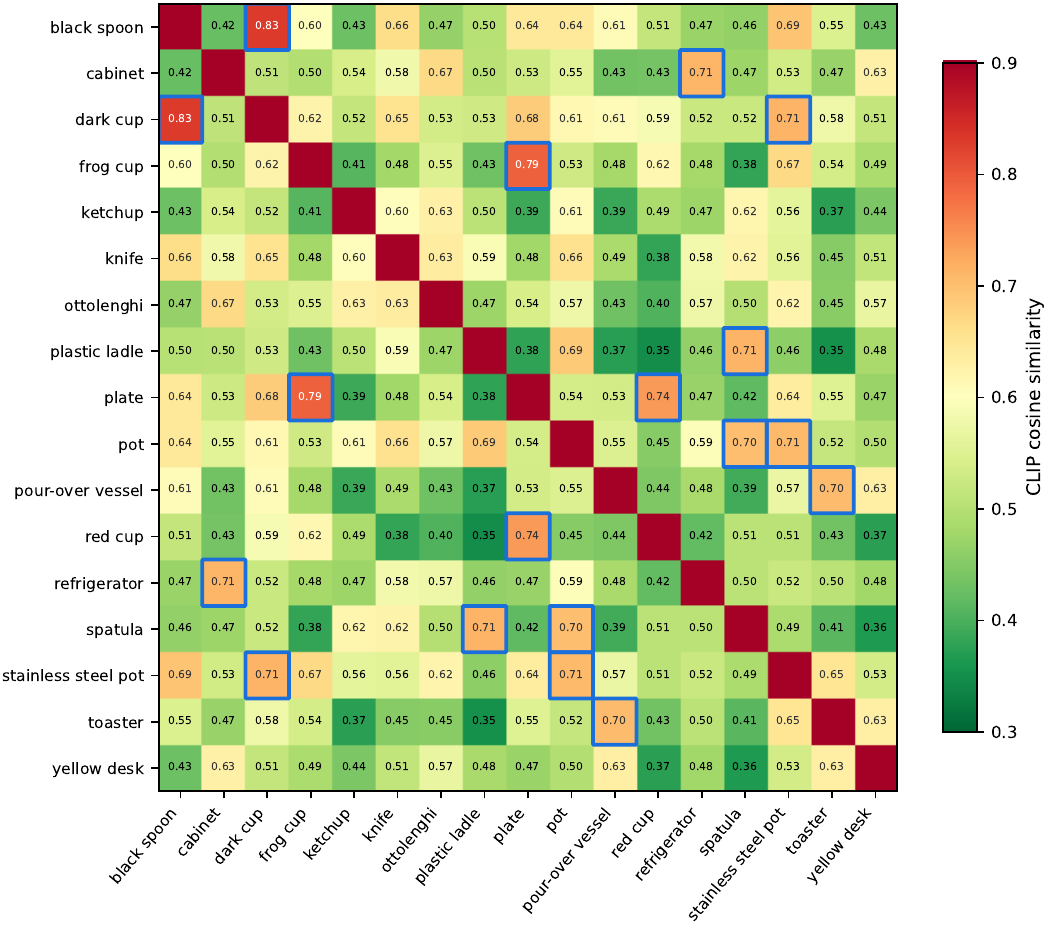}
    \subcaption{\textit{waldo\_kitchen}: localized confusion pairs.}
    \label{fig:sim-waldo}
  \end{minipage}
  \caption{\textbf{Anchor-space ambiguity patterns motivating Gate~3.} Blue borders mark anchor pairs with similarity ${>}0.70$. The two scenes exhibit different ambiguity structures, but both make naive relabeling unreliable and motivate the margin-based safeguard in GSFL.}
  \label{fig:clip-sim-comparison}
\end{figure*}

\begin{table}[ht]
\centering
\caption{Quantitative 3D semantic segmentation results on Mip-NeRF360 sparse scenes, reported as mean IoU (\%).}
\label{tab:seg3}
\resizebox{1\linewidth}{!}{%
\begin{tabular}{lccccc}
\toprule
\textbf{Method} & room & counter & garden & bonsai & Overall \\
\midrule
GS--Grouping\cite{ye2024gaussiangrouping} & $47.7$ & $40.4$ & $54.1$ & $49.2$ & $54.4$ \\
GOI\cite{qu2024goi} & $60.3$ & $46.6$ & $59.8$ & $67.3$ & $58.5$ \\
OpenGaussian\cite{wu2024opengaussian} & $56.6$ & $39.4$ & $32.2$ & $44.1$ & $48.0$ \\
LangSplat\cite{qin2024langsplat} & $53.2$ & $68.8$ & $51.9$ & $55.4$ & $57.3$ \\
LangSplat--v2\cite{li2026langsplatv2} & $64.3$ & $75.1$ & $65.0$ & $73.1$ & $69.4$ \\ \midrule
Ours & $\mathbf{65.2}$ & $\mathbf{80.9}$ & $\mathbf{66.3}$ & $\mathbf{89.3}$ & $\mathbf{75.4}$ \\
\bottomrule
\end{tabular}%
}
\end{table}
\noindent\textbf{Ablation Study.} Table~\ref{tab:abl-seg} evaluates the contribution of each gate in GSFL on the \textit{ramen} scene. The SAD-GS model achieves the best performance at $\mathbf{77.9\%}$ mIoU, outperforming SAD-only ($72.5\%$). Removing any gate lowers performance. The largest drop comes from removing Gate~2, which reduces mIoU to $41.5\%$ and shows that the absolute acceptance threshold ($\tau_{\text{acc}}$) is the main barrier against incorrect relabeling. Removing Gate~3 also causes a clear drop, confirming that the margin constraint ($\delta$) is needed when multiple anchors remain plausible. Fig.~\ref{fig:clip-sim-comparison} clarifies this role: ambiguity in \textit{figurines} is global, whereas ambiguity in \textit{waldo\_kitchen} is concentrated in a few local confusion pairs. Gate~3 does not change this anchor-space structure; it makes relabeling safer by requiring a clear margin over the runner-up.

\begin{figure*}[ht]
    \centering
    \begin{minipage}[t]{0.48\linewidth}
        \centering
        \includegraphics[width=\linewidth]{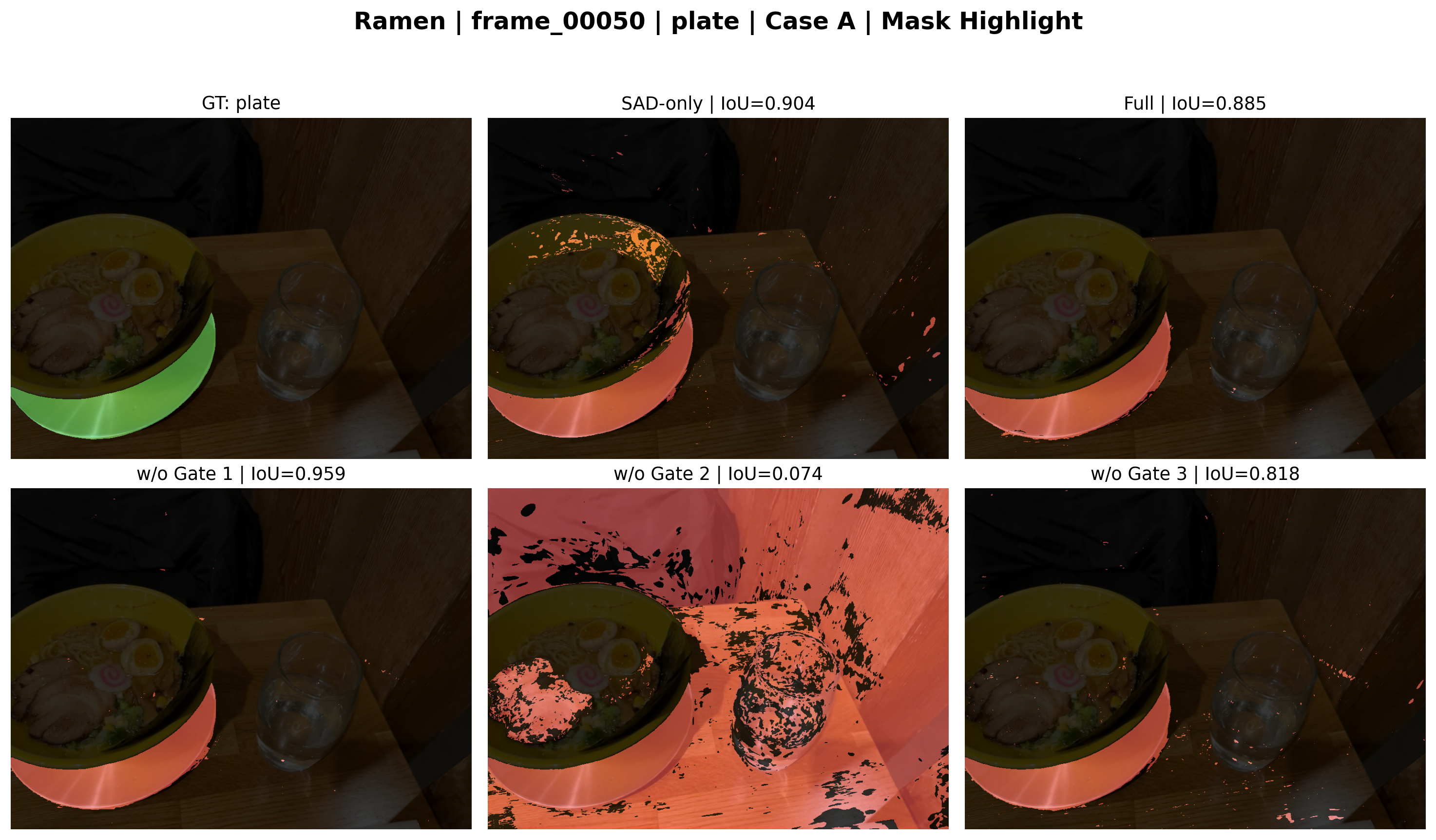}
    \end{minipage}\hfill
    \begin{minipage}[t]{0.48\linewidth}
        \centering
        \includegraphics[width=\linewidth]{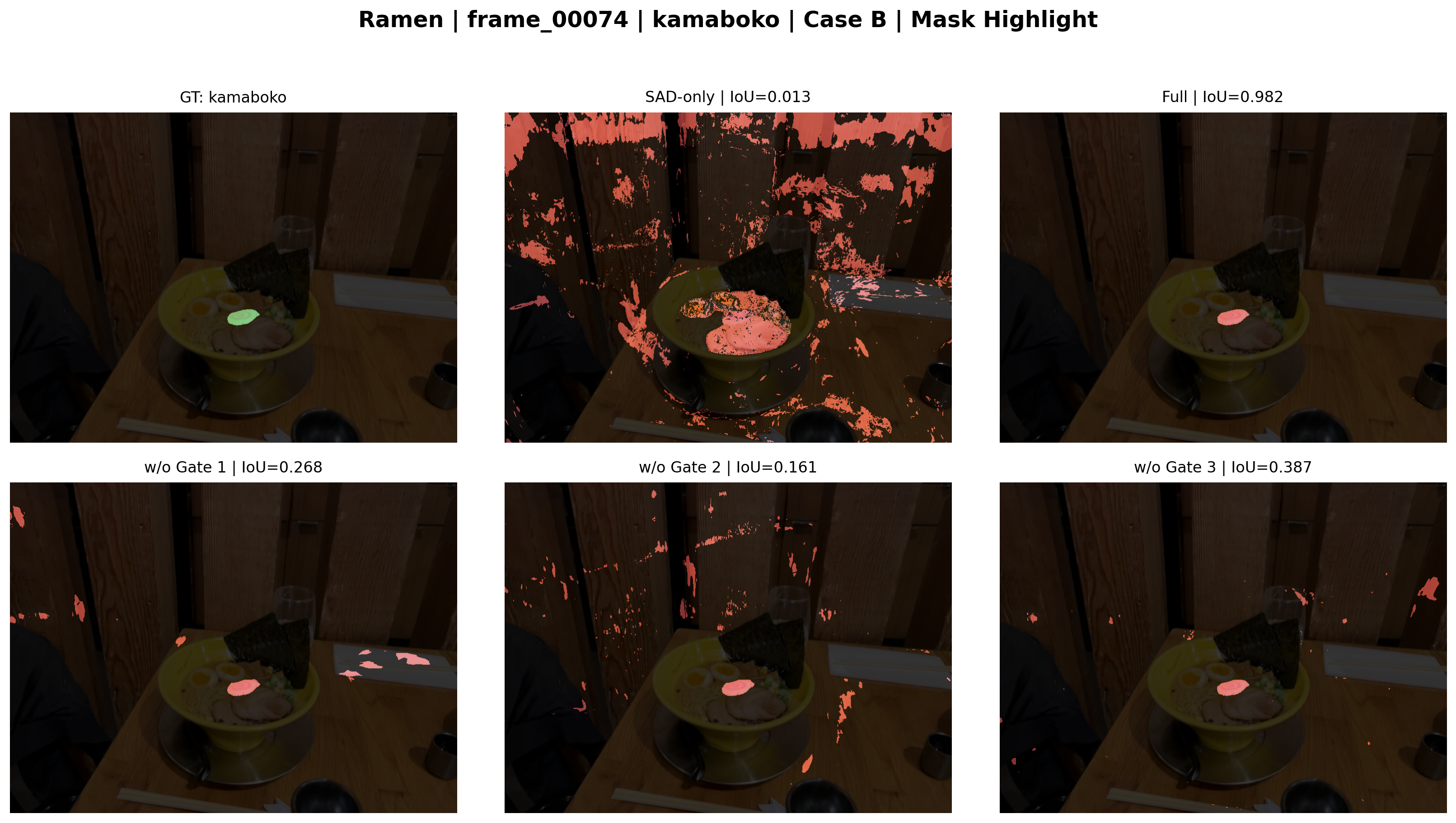}
    \end{minipage}
   \caption{\textbf{Visual comparison of gate-removal ablations on \textit{ramen}.} The full three-gate design yields the cleanest masks, while removing any gate degrades pseudo-label refinement.}
   \label{fig:gate-cases}
\end{figure*}

Fig.~\ref{fig:gate-cases} provides qualitative evidence for the same conclusion. In Case~A (\textit{plate}), SAD-only shows substantial semantic leakage inside the bowl, whereas the SAD-GS model produces a much cleaner mask. In Case~B (\textit{kamaboko}), SAD-only fails to recover the target, while the full model succeeds. Removing any gate degrades the result, and removing Gate~2 causes the most severe failure. These examples show that the three gates work together to make pseudo-label refinement robust.

\begin{table}[ht]
    \centering
    \caption{Gate-removal ablation on \textit{ramen}. The full three-gate design remains the strongest overall setting, while removing Gate~2 produces the largest regression.}
    \label{tab:abl-seg}
    \resizebox{1\linewidth}{!}{
    \begin{tabular}{lcccc}
    \toprule
    \textbf{Variant} & \textbf{mIoU (\%)} & $\Delta$ vs.\ Full & $T$ & $f_{\text{fg}}$ \\
    \midrule
    SAD-only & 72.5 & $-5.4$ & 0.13 & 0.50 \\
    SAD + PLR w/o Gate~1 & 72.8 & $-5.1$ & 0.07 & 0.65 \\
    SAD + PLR w/o Gate~2 & 41.5 & $-36.4$ & 0.07 & 0.65 \\
    SAD + PLR w/o Gate~3 & 69.0 & $-8.9$ & 0.10 & 0.45 \\ \midrule
    \textbf{SAD-GS} & \textbf{77.9} & \textbf{--} & \textbf{0.11} & \textbf{0.65}  \\
    \bottomrule
    \end{tabular}}
\end{table}

\begin{figure}[ht]
    \centering
    \includegraphics[width=\linewidth]{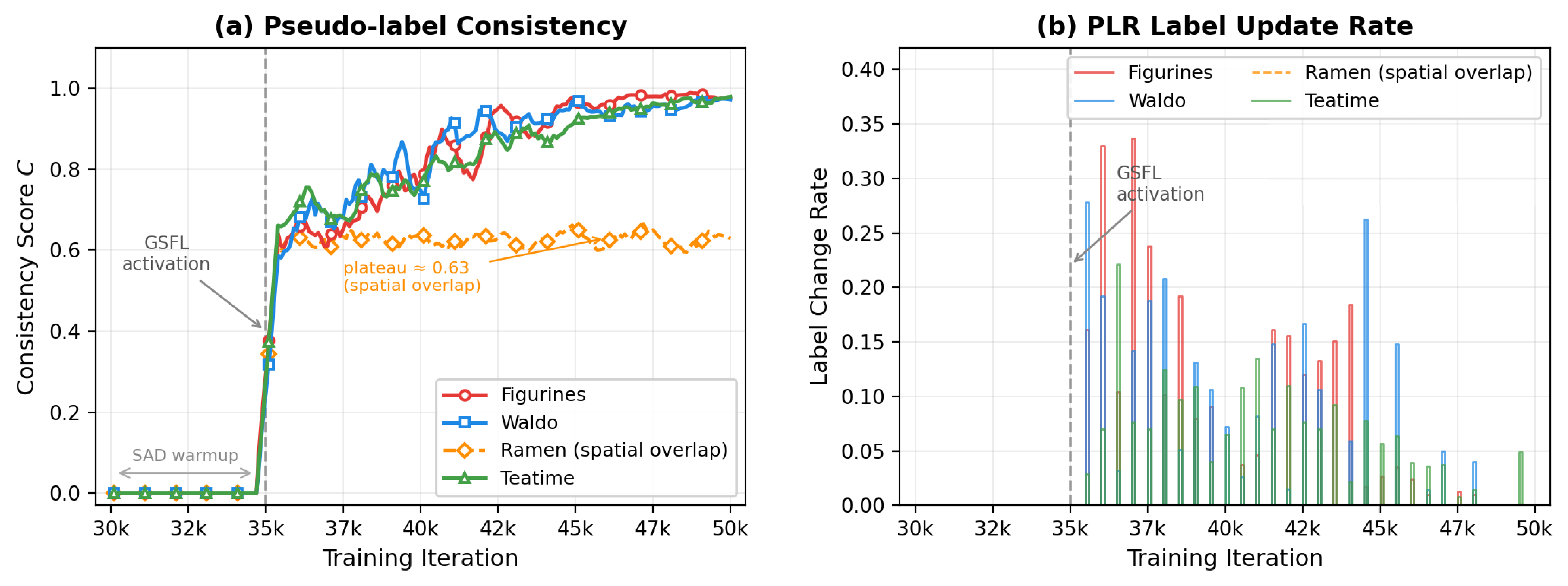}
     \caption{\textbf{GSFL convergence on representative scenes.} After activation, geo-semantic consistency rises while label updates are progressively suppressed as training stabilizes.}
    \label{fig:gsfl-convergence}
\end{figure}

\subsection{Closed-Loop Dynamics and Sensitivity Analysis}
\noindent\textbf{Feedback Convergence Tracking.} 
Fig.~\ref{fig:gsfl-convergence} shows that GSFL behaves as a closed-loop refinement process after activation at iteration 35{,}000. Across \textit{figurines}, \textit{waldo}, and \textit{teatime}, the consistency score $C$ rises above $0.96$ while the label-update rate decays toward zero, indicating progressively self-consistent pseudo-label correction. On \textit{ramen}, $C$ stays near $0.63$ and the update rate remains almost zero even after activation, showing that GSFL suppresses further relabeling when heavy spatial overlap leaves the ambiguity unresolved. This supports the role of GSFL as a stabilizing feedback mechanism rather than an always-on source of label changes.

\begin{figure}
    \centering
    \includegraphics[width=1\linewidth]{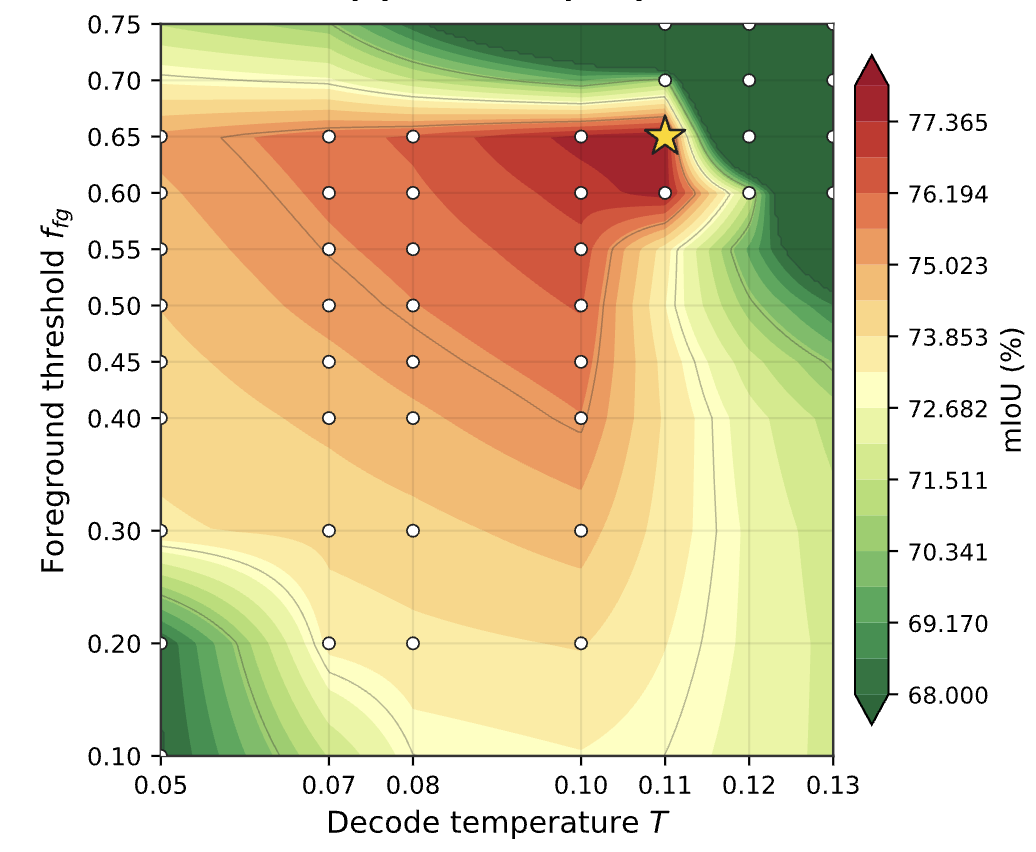}
\caption{\textbf{Decode robustness of SAD-GS on \textit{ramen}.} SAD-GS maintains strong segmentation performance across a broad range of decoding temperature $T$ and foreground threshold $f_{\text{fg}}$, showing that its gains are not tied to a narrow threshold choice.}
    \label{fig-contour}
\end{figure}

\textbf{Decoding Robustness Analysis.} Fig.~\ref{fig-contour} evaluates the performance stability of \textbf{SAD-GS} under varying inference configurations. We sweep the decoding temperature $T$ and foreground threshold $f_{\text{fg}}$ over a broad range on the \textit{ramen} scene. The full model maintains high mIoU scores across a wide parameter region rather than relying on a narrow operating point. This decoding robustness indicates that the performance gains stem from a more reliable 3D semantic field rather than sensitive post-hoc threshold tuning. Performance drops only under extreme decoding settings where the relevancy maps become either over-saturated or overly conservative.

\section{Conclusion and Limitations.}
In this paper, we have presented SAD-GS for robust open-vocabulary 3D semantic Gaussian splatting, designed to overcome the supervision reliability limitations of previous language fields. To achieve this, SAD stabilizes semantic identities using consensus text anchors, while GSFL refines noisy tracker labels through geometry-aware feedback, significantly improving both localization and segmentation across multiple benchmarks. Despite its strong performance, \textbf{SAD-GS} still has several limitations. Its correction gains can saturate under severe spatial overlap or inside complex container where pure text anchoring provides an incomplete supervisory signal. Furthermore, when many semantic categories cluster closely in the CLIP space, the fixed global margin threshold becomes less effective. Future work could address these by incorporating spatio-temporal priors or developing class-adaptive local decision boundaries to handle persistent ambiguity.

\clearpage
\newpage

{
    \small
    \bibliographystyle{ieeenat_fullname}
    \bibliography{software}
}

\end{document}